# The Dynamic of Body and Brain Co-Evolution

**Paolo Pagliuca and Stefano Nolfi**
Institute of Cognitive Sciences and Technologies,
National Research Council, Italy
paolo.pagliuca@istc.cnr.it, stefano.nolfi@cnr.it

**Abstract**
We introduce a method that permits to co-evolve the body and the control properties of robots. It can be used to adapt the morphological traits of robots with a hand-designed morphological bauplan or to evolve the morphological bauplan as well. Our results indicate that robots with co-adapted body and control traits outperform robots with fixed hand-designed morphologies. Interestingly, the advantage is not due to the selection of better morphologies but rather to the mutual scaffolding process that results from the possibility to co-adapt the morphological traits to the control traits and vice versa. Our results also demonstrate that morphological variations do not necessarily have destructive effects on robot skills.

## 1. Introduction

The ability of a robot to successfully achieve its goals in its environment depends both on its physical structure and on its proficiency at control, which are inherently coupled (Pfeifer & Bongard, 2006). This implies that both the characteristics of the robot's controller and of the robot's body should be adapted to maximize performance. This is the objective of the research area that studies the co-evolution of body and brain, a research field initiated by the seminal work of Karl Sims (1994).

The objective of these studies is challenging. A first difficulty derives from the need to identify a suitable set of elementary elements that can be combined to generate agents with highly variable bodies. A second difficulty comes from the need to adapt agents formed by heterogeneous and interdependent elements (e.g. body parts, sensors, actuators, neurons, neural connections). A third difficulty is caused by the need to carry on realistic and computationally expensive simulations or expensive and time-consuming experiments in hardware. Finally, a fourth problem is caused by the fact that each study investigates a different problem and operates with robots that differ qualitatively from related studies, making the comparison of obtained results difficult. As a consequence of all these difficulties, the progress achieved in the area over the last 25 years was limited, despite the increase of computer processing power (Cheney et al., 2016) .

In this paper we investigate whether the co-evolution of the morphological and control properties permits to achieve higher performance with respect to a control condition in which the morphology of the agent is hand-designed and fixed. We demonstrate that this is indeed the case in the context of robot locomotor problems (Todorov, Erez & Tassa, 2012; Coumans & Bay, 2016), a widely used benchmark for continuous control optimization problems. Interestingly, the obtained results show that the advantage gained by co-evolving morphology and control is due primarily to the possibility to co-adapt the two properties, and only secondarily to the possibility to discover optimized morphology. We show that the adaptation of both morphological and control characteristics continues in the long-term without producing a premature convergence on suboptimal solutions. Finally, we show how the method proposed can be extended to evolve both the overall bauplan and the detailed characteristics of the robots' body.

In Section 2 we discuss the main alternative methods proposed to co-evolve body and brain. In Section 3 we describe the experimental setting chosen and the motivations behind it. In Section 4 we report the obtained results. In Section 5 we describe the experiments in which the overall bauplan of the robot's body is subjected to evolution. Finally, in Section 6 we discuss the implication of our results.

## 2. Co-evolving body and brain

In this section we briefly review the alternative methods proposed to co-evolve body and brain, the relative advantages and drawbacks, and the motivation behind the selection of our method. The review does not aim to be exhaustive. It rather aims to identify the qualitative differences between the proposed methods and their implications.

In his seminal work, Sims (1994) evolved neuro-controlled creatures formed by 3D rigid parts assembled through fixed and actuated joints. The genotypes of evolving creatures encode the developmental instruction for growing a creature represented in a directed graph formed by nodes and connections (that can be recurrent). The generation of the phenotype starts from a root node, which contains the instruction for synthesizing an initial body element, and continues with the generation of the body elements encoded in connected nodes. Body elements include: (i) sensors (join angle, contact, and light sensors) with associated sensory neurons, (ii) actuators controlling the DOFs of the joint with associated motor neurons, and (iii) internal nodes/neurons. The nodes of the directed graph contain the parameters that specify: (i) the dimensions of the body element, (ii) the type of the joint connecting it to the preceding element, (iii) the function computed by each local neuron (selected among a list of available functions), and (iv) the connection and the connection weights with neurons located in the same element and in other elements. The connections of the directed graph contain parameters that specify the displacement, orientation, scale and reflection of the new element with respect to its parent element.

Later related works adopted a similar method but relied on a smaller repertoire of elementary elements and on a simpler genotype-to-phenotype encoding. In the case of Lipson & Pollack (2020), for example, the evolving robots are composed of cylinders of variable length, actuators, and neurons. The actuators control the telescopic joints that determine the length of the cylinders. The genotype of the evolving robots is formed by a list of tuples that encode: (i) vertices (i.e. points in a 3D Euclidean space), (ii) cylinders with associated parameters (i.e. index of starting and ending vertex, length, and stiffness of the motorized joint), (iii) neurons with associated parameters (i.e. connection weights to all other neurons), and (iv) actuators with associated parameters (i.e. index of the cylinder, index of the motor neuron controlling the motor, and extension range).

Embryogenetic approaches (Dellaert & Beer, 1996; Eggenberger Hotz, 1997; Joachimczak, Suzuki & Arita, 2016) are inspired more directly by the developmental process that characterizes natural multicellular organisms. One of the most interesting models of this class (Joachimczak, Suzuki & Arita, 2016) is based on 2D spherical cells, connected through elastic springs, that grow and differentiate from a single initial cell. The initial cell divides in two cells that then eventually divide in four cells and so on. The fate of cells is determined by a simple abstracted genetic regulatory network implemented in a feed-forward neural network. This genetic regulatory network receives as input the Cartesian coordinates of the cell and the signals received by nearby cells. The network produces as output two binary values that determine: (i) whether the cell divides or not, (ii) whether the cell dies or not, (iii) the relative positions of the new cells originating from the division, (iv) the signals transmitted to nearby cells, (v) the frequency of the contraction of the spring attached to the cell, and (vi) the phase shift of the contraction.

In a related model proposed by Cheney, Clune & Lipson (2014), instead, the agents are formed by 3D cubic cells of different types arranged within a 3D grid of 10x10x10 voxels (see also Corucci et al., 2018). The cells belong to one of the four following types: cells made of rigid material, cells made of soft material, cells made of soft material that contract and expand periodically, and cell made of soft material that expand and contract periodically. The presence/absence of the cell and the type of the cell is determined by a neural network that receives as input the Cartesian coordinate of each voxel and determines as output the presence/absence of the cell and the type of the cell.

Eventually, cells can also emit signals that diffuse over space, which can regulates the contracting/expanding phase of nearby cells.

Finally, another approach consists in adapting agents formed by a predetermined number of body elements with configurable morphological properties (Ha, 2019; Nygaard et al., 2020). In the case of Nygaard et al. (2020), for example, the agent consists of a quadruped robot trained for the ability to walk on different terrains and the configurable properties are the length of the femur and of the tibia of the legs.

We can assume that the ideal approach to body and brain co-evolution should be simple, should have a high expressive power, and should allow the generation of evolvable solutions. In other words, the ideal approach should postulate a simple set of elementary elements (analogous to the amino acids in natural organisms), should be capable to generate a wide spectrum of phenotypes (analogous to the huge number of species produced by natural evolution), and should permit the generation of agents with a sufficient propensity to improve as a result of genetic variations.

The proposed methods vary substantially with respect to these key properties. In particular, the model proposed by Sims (1994) has the greatest expressive power but lacks simplicity. The expressive power is granted by the usage of several classes of elementary components (body elements, neurons, proprioceptors, exteroceptors, actuators) that are highly configurable. The undesirable complexity of the model derives from the need to choose the properties and the variation ranges of the long list of elementary components used. The models proposed by Joachimczak, Suzuki & Arita (2016) and by Cheney, Clune & Lipson (2014) are much simpler but do not permit the evolution of agents with sensing capabilities and with structured control architectures.

Measuring and comparing the evolvability of methods that differ so widely among themselves is difficult. Most of the proposed methods show the ability to generate clever solutions and qualitatively different solutions in different evolutionary runs. On the other hand, the chance of producing improvements generally decreases quickly after few hundreds of generations. In particular, as pointed out by Joachimczak, Suzuki & Arita (2016) and Cheney et al. (2016), the morphology of the evolving agents converges on a stable configuration after few hundreds of generations. This premature convergence prevents the possibility to generate better and better solutions.

## 3. Method

To study the dynamic of body and brain adaptation we used a direct encoding method in which the body of the agent is formed by a fixed number of body elements with genetically encoded properties. We choose this approach since it allows to compare hand-designed and evolved morphologies and since it can be used in combination with state-of-the-art evolutionary algorithms (Salimans et al., 2017, see also Pagliuca, Milano & Nolfi, 2020) that operate on continuous parameters only. We applied this method to evolve the morphological properties of agents that have a fixed hand-designed bauplan (as in the case of Ha, 2019) and to evolve also the morphological bauplan of the agents. Moreover, we choose this approach since it permits to verify whether the concurrent co-adaptation of morphological and control features is beneficial or disadvantageous. The fact that the parameters encoding the morphological and control features are clearly separated, in fact, permits to study the implication of concurrently adapting the two type of parameters or adapting one set at a time. The potential advantage of adapting the morphological and control parameters in parallel is that this can produce a mutual scaffolding process in which the characteristics of the morphology are adapted to the current characteristics of the controller and the characteristic of the controller are adapted to the current characteristics of the morphology. The potential disadvantage of adapting the two types of feature in parallel is that morphological variations can have destructive effect on robots skills since the function performed by the controller depends crucially on the robot/environmental interaction that is mediated by the morphological characteristics of the agent (Cheney et al, 2016).

As a testbed we used by the Pybullet locomotor problems (Coumans & Bai, 2016). These environments represent a free and more realistic implementation of the MuJoCo locomotor problems designed by Todorov, Erez & Tassa (2012), which constitutes a widely used benchmark for continuous control domains. Pybullet locomotors are composed by several cylindrical body parts connected through hinge joints. More specifically, in our experiments we consider the Walker2D and Halfcheetah robots. Both robots include two legs formed by a femur, a tibia and a foot. In the case of the Walker2D robot, the two legs are attached to the bottom part of the torso element that extends vertically (Figure 1, left). In the case of the Halfcheetah, instead, the two legs are attached to the frontal and rear end of the torso element that extends horizontally. Moreover, the torso includes a head segment attached on the frontal side (Figure 1, right). The body parts are attached to the torso or to the previous segment through actuated hinge joints with the exception of the head element of the Halfcheetah, that is attached through a fixed joint.

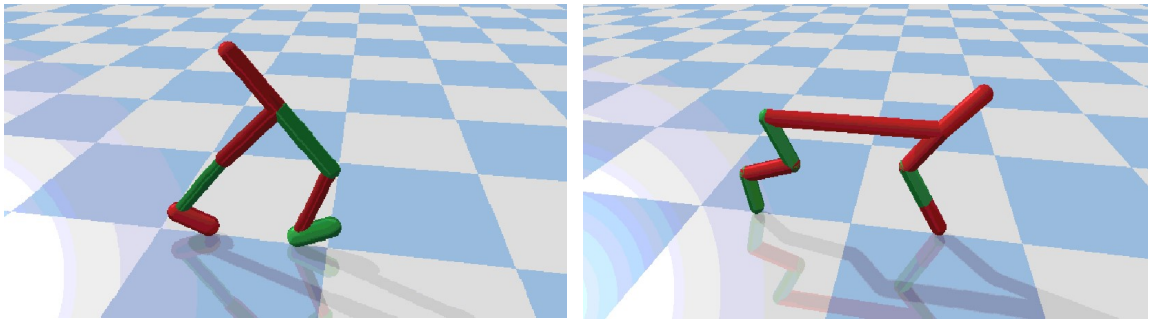

Figure 1. The Walker2D and Halfcheetah robots (left and right respectively).

In the experiments reported in Section 4, the general bauplan of the Walker2D and Halfcheetah robots is maintained, i.e. the number elements, the number of joints, and the default angle of the joints of the original hand-designed robots are preserved and are not subjected to variation. However, the length and the diameters of the segments (and consequently the mass of the elements), and the offset of clockwise and anti-clockwise limits of the joints are allowed to vary. These characteristics are encoded as free parameters and evolved together with the connection weights of the neural network controller of the robot. In the case of the Walker2D, the morphological parameters of the two legs are shared.

Overall, the parameters include 14 and 34 morphological traits in the case of the Walker2D and Halfcheetah, respectively. Each parameter is squashed in the range [-1.0, 1.0] through the tanh function and then normalized in the range [-20%, +20%] where 0.0 corresponds to the hand-designed value of the parameter specified in PyBullet and -20% and 20% correspond to the minimum and maximum possible variation of the parameter value.

The robots are rewarded at each time step on the basis of their velocity in m/s toward the target destination. In the case of the experiments with the Halfcheetah, the robots are also penalized every step with -0.1 for each joint that is in a limit position (for an analysis of the utility of this additional fitness component, see Pagliuca, Milano, and Nolfi, 2020). Moreover, to avoid the evolution of unrealistic behaviors that exploit the limits of the simulator, the agents are also strongly penalized when the velocity of their joints exceeds 5 m/s. The agents are evaluated for 1 episode lasting 1000 steps. At the beginning of the episode the agent is situated slightly above the ground. To promote the evolution of robust solutions, the initial position of the joints is randomly perturbed by altering the angular position of each joint of a random value in the range [-0.1, 0.1] rad, selected with a uniform distribution. Evaluation episodes are terminated prematurely when the agents fall down or tip over (i.e., when the inclination of the torso exceeds 1.0 rad, or when the elevation of the torso is below 0.8 and 0.3 in the case of the Walker2D and Halfcheetah, respectively, or when the head, the torso, the femurs, or the tibias enter in contact with the ground in the case of the Halfcheetah).

The controller of the robot is constituted by a feedforward neural network with 22 and 26 sensory neurons in the case of the Walker2D and Halfcheetah respectively, 50 internal neurons, and 6 motor neurons. The sensory neurons encode the orientation and the velocity of the robot, the relative orientation of the target destination, the position and the velocity of the six joints, and the state of the two contact sensors located on the feet. The motor neurons encode the intensity and the direction of the torque applied by the 6 motors controlling the 6 corresponding joints. Forces pushing joints toward their limits are ignored for joints that already reached their limits. The internal and output neurons are updated by using the tanh and linear activation functions, respectively. The state of the motor neurons is perturbed each step with the addition of Gaussian noise with mean 0.0 and standard deviation 0.01. The 1456 and 1656 connection weights and biases of the neural network of the Walker2D and Halfcheetah are encoded in free parameters and evolved together with the morphological parameters.

The morphological and control parameters are evolved through the OpenAI evolutionary strategy (Salimans et al., 2017), a state-of-the-art method for continuous control optimization problems (see also Pagliuca, Milano, and Nolfi, 2020). The genotype of the evolving robots is formed by 1456 control parameters and 14 morphological parameters in the case of the Walker2D and 1656 control parameters and 34 morphological parameters in the case of the Halfcheetah. In control experiments in which the morphology of the robots remains fixed, the genotype includes 1456 and 1656 parameters only, in the case of the Walker2D and Halfcheetah respectively. As in the experiments performed by (Salimans et al., 2017), observation states are normalized by using the virtual batch normalization (Salimans et al., 2016, 2017) and the parameters of the genotype are normalized by using L1 weight decay (Ng, 2004). The weight decay is applied to the connection weights of the neural network and to the morphological parameters. In the case of the neural network, the usage of weight decay improves generalization and preserves the evolvability of the network (see also Pagliuca, Milano and Nolfi, 2020). In the case of the body morphology, the usage of weight decay biases the search toward the default value of the parameters, thus reducing neutral drift. The evolutionary process is continued until the total number of steps performed exceeds $10^8$.

The source code that can be used to replicate experiments is available at https://github.com/PaoloP84/BodyBrainCoevolution.

## 4. Results

To verify the advantage of co-evolving the morphological features, we run a set of experiments with the model described in Section 3 (co-evolving condition) and an additional set of control experiments in which the morphological characteristics are hand-designed and fixed (fixed condition).

The obtained results indicate that the co-evolved condition produces significantly better results than the fixed condition (Figure 2, Mann-Whitney U test, p-value < 0.05).

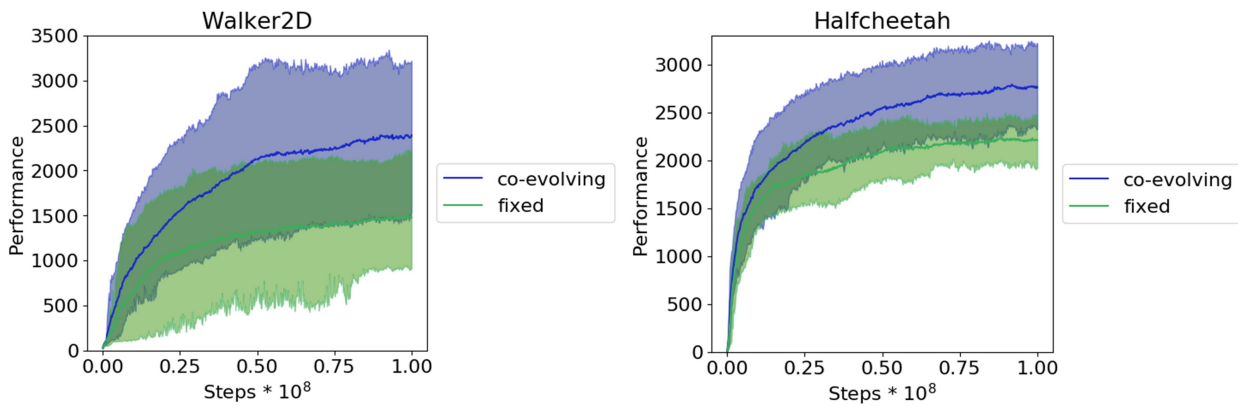

Figure 2. Performance of the best agents during the course of the evolutionary process for the experiments performed in the co-evolved and fixed conditions. The left and right graph show the results obtained on the Walker2D and Halfcheetah problems, respectively. Performance refers to the average fitness obtained by post-evaluating the best individual of each generation for 3 evaluation episodes. Data collected over 20 replications for each condition. Shadow areas indicate 90% bootstrapped confidence intervals of the performance across 20 replications per experiment.

To verify whether the morphology of the evolving agents keeps changing during the entire evolutionary process or converges over a structure that remains constant from a certain phase on, we measured the average absolute difference of the control and morphological parameters between the current best agent and the previous best agent, every $10^5$ evaluation steps (Figure 3). As can be seen, after an initial burst of variation, both the morphological and control parameters keep changing during the entire evolutionary process at an almost constant rate. These data thus indicate that both the control and morphological characteristics of the agents keep changing during the entire evolutionary process.

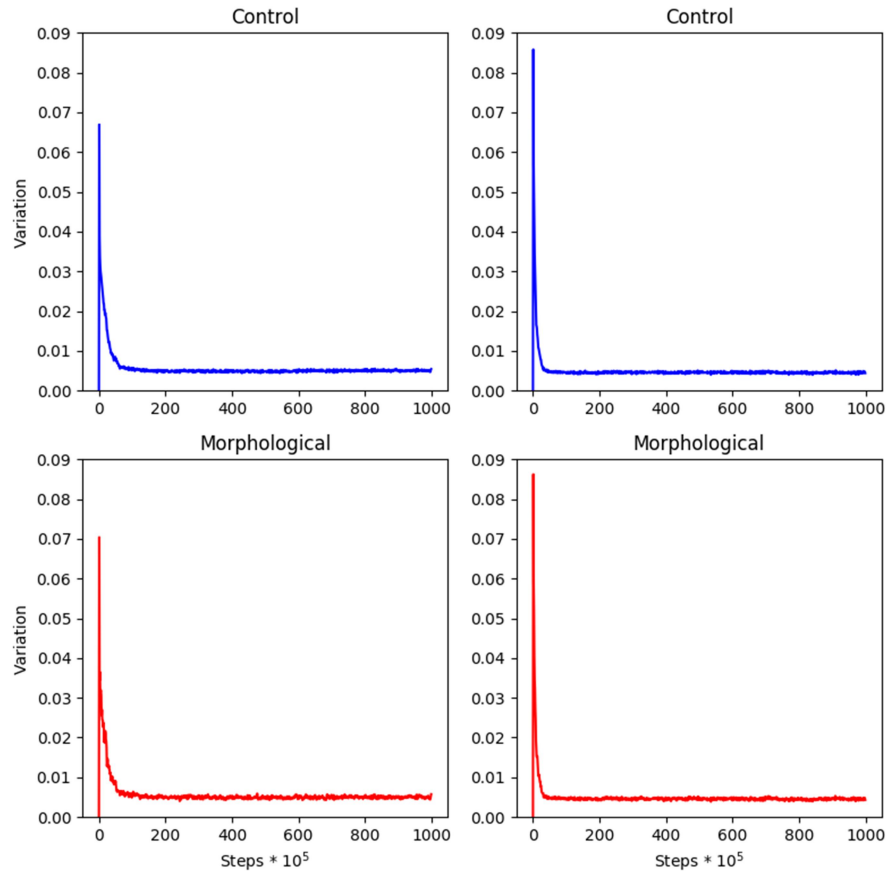

Figure 3. Average variation of the control and morphological parameters (top and bottom, respectively) in the case of the experiment performed with the walker2D and Halfcheetah problems (left and right, respectively). Data collected over 20 replications for each condition.

To verify to what extent the performance gain achieved by the co-adaptation experimental condition depends on the superiority of the evolved morphologies with respect to the hand-designed morphologies and/or on the possibility to adapt the morphological traits to the control traits and vice versa, we ran a new set of evolutionary experiments in which the morphological traits of the robots were set equal to those of the best robots evolved in the co-evolving condition and were maintained constant during the evolutionary process (pre-evolved experimental condition). The results of this and the other conditions are reported in Figure 4.

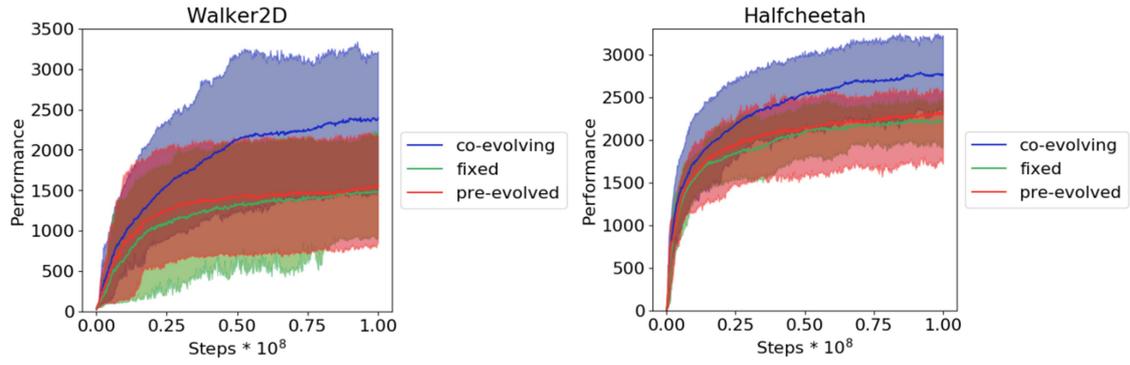

Figure 4. Performance of the best agents obtained during the course of the evolutionary process in experiments performed in the fixed, co-evolving, and pre-evolved conditions. The left and right graph, show the results obtained on the Walker2D and Halfcheetah problems. Performance refers to the average fitness obtained by post-evaluating the best individual of each generation for 3 episodes. Shadow areas indicate 90% bootstrapped confidence intervals of the performance across 20 replications per experiment.

The agents evolved in the co-evolved conditions outperform those evolved in the pre-evolved conditions (Figure 4, Mann-Whitney U test with Bonferroni correction, p-value < 0.05). The performance of the agents of the pre-evolved condition does not differ significantly from the performance of the agents of the fixed condition (Figure 4, Mann-Whitney U test with Bonferroni correction, p-value > 0.05). Overall, these data indicate that the co-evolution of control and morphological traits permit to discover better solutions than experiments based on a fixed hand-designed morphology. Moreover, these results indicate that the advantage provided by the co-evolution of control and morphological traits in these experiments is not due to the possibility to discover better morphologies, but rather to the possibility to adapt the morphological traits to the control traits and vice versa.

## 5. Evolving the general bauplan and the detailed characteristics of the agents' body

In this section we report an additional set of experiments involving agents formed by a collection of identical elements initially arranged in a uniform morphology. As in the case of the experiments reported above, the agents are constituted by cylinders attached to other cylinders through actuated hinge joints and inclined of a certain default angle. Moreover, as in the case of the previous experiments, the length, the radius, and the clockwise and anticlockwise limits of the joints are encoded in morphological parameters and are thus variable. However, in this case each element is attached to the endpoint of the previous element. Moreover, the default inclination of each element with respect to the previous element is encoded in morphological parameters and allowed to vary (see Table 1). This produces uniform linear structures (see Figure 5) that can then vary during the course of the evolutionary process by eventually producing structure similar to those of hand-designed agents. The identification of a suitable morphological bauplan is demanded to the evolutionary process together with the identification of the detailed morphological and control traits.

The objective of these new sets of experiments is thus that to verify whether evolution can successfully "sculpt" a suitable morphological bauplan and the detailed characteristics of the body elements, through the accumulation of small morphological variations, or not.

Table 1. The default value and the variation range of the properties of body elements

|  | Default Value | Min-Variation-Range | Max-Variation- |
|---|---|---|---|
| length of segments | 0.5 m | -0.25 m | 0.25 m |
| radius of segments | 0.05 m | -0.025 m | 0.025 m |
| default angle of joints | 0 degrees | -135 degrees | +135 degrees |

| clockwise limit of joints | 0 degrees | -180 degrees | +180 degrees |
| anticlockwise limit of joints | 0 degrees | -180 degrees | +180 degrees |

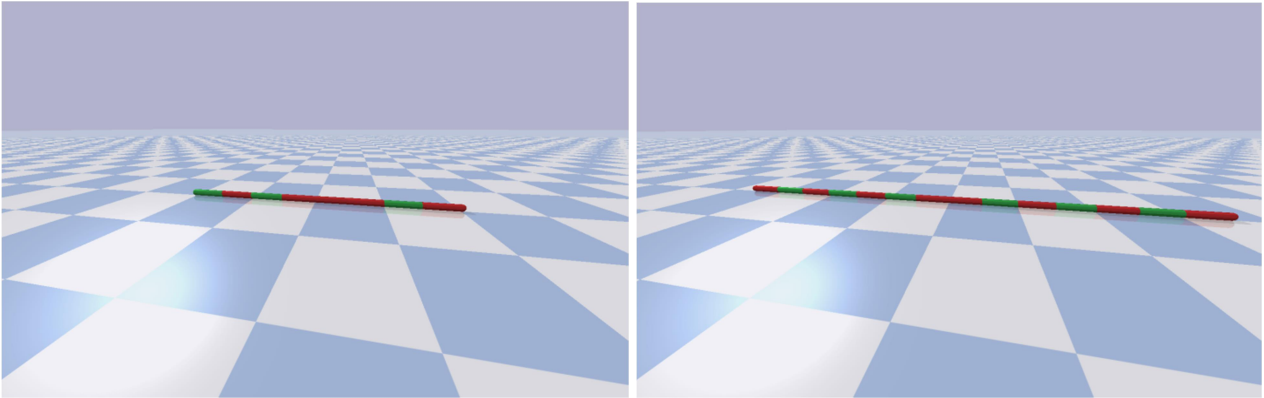

Figure 5. The initial topology of the evolving agents is constituted by identical elements arranged in a linear structure. Left: an agent formed by 7 elements. Right: an agent formed by 13 elements.

As in the case of the previous experiments, the robots are rewarded primarily on the basis of their velocity in m/s toward the target destination. To channel the evolutionary process toward effective locomotion behaviors, the agents are also penalized with 0.1 for every segment in contact with the floor and with 0.001 (0.002 in the case of an agent made of 13 elements) for every newton of energy used to control the joints. Finally, as in the case of the experiments reported in the previous section, the robots are penalized when their body parts and/or joints exceed a velocity of 5 m/s.

We run two series of experiments with agents composed of 7 and 13 body elements (see Figure 5). In the former case, the policy network includes 27 sensory neurons, 50 internal neurons, and 6 motor neurons. The genotype of the evolving agents includes 32 morphological parameters and 1706 control parameters (connection weights). In the latter case the policy network includes 45 sensory neurons, 50 internal neurons, and 12 motor neurons. The genotype of the evolving agents includes 62 morphological parameters and 2912 control parameters.

Figure 6 shows the performance across generations for the two series of experiments (see co-evolving condition).

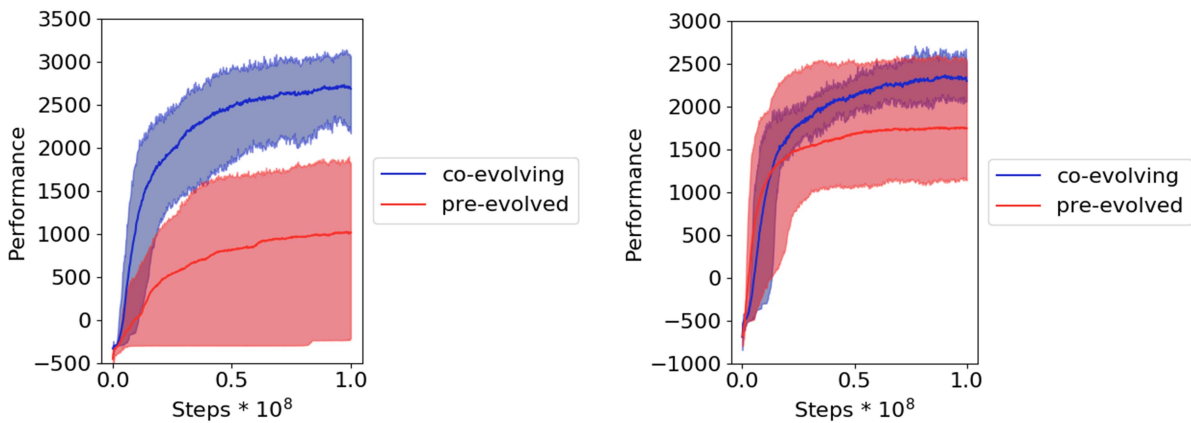

Figure 6. Performance of the best solution obtained during the course of the evolutionary process in experiments performed with agents formed by 7 and 13 elements, left and right graph respectively. The two curves display the results obtained in the co-evolving and pre-evolved experimental conditions. Performance refer to the average fitness obtained by post-evaluating the best individual of each generation for 3 evaluation episodes. Shadow areas indicate 90% bootstrapped confidence intervals of the performance across 20 replications per experiment.

The fact that the evolving agents manage to achieve performance similar to those obtained in the experiments reported in section 4 shows that the lack of a hand-designed bauplan does not prevent the evolution of suitable solutions. This is also confirmed by the visual inspection of the evolved agents. As shown in Figure 7, in fact, the evolving agents discover articulated morphologies that include supporting elements, walking legs, and balancing elements (see also the video included in the appendix).

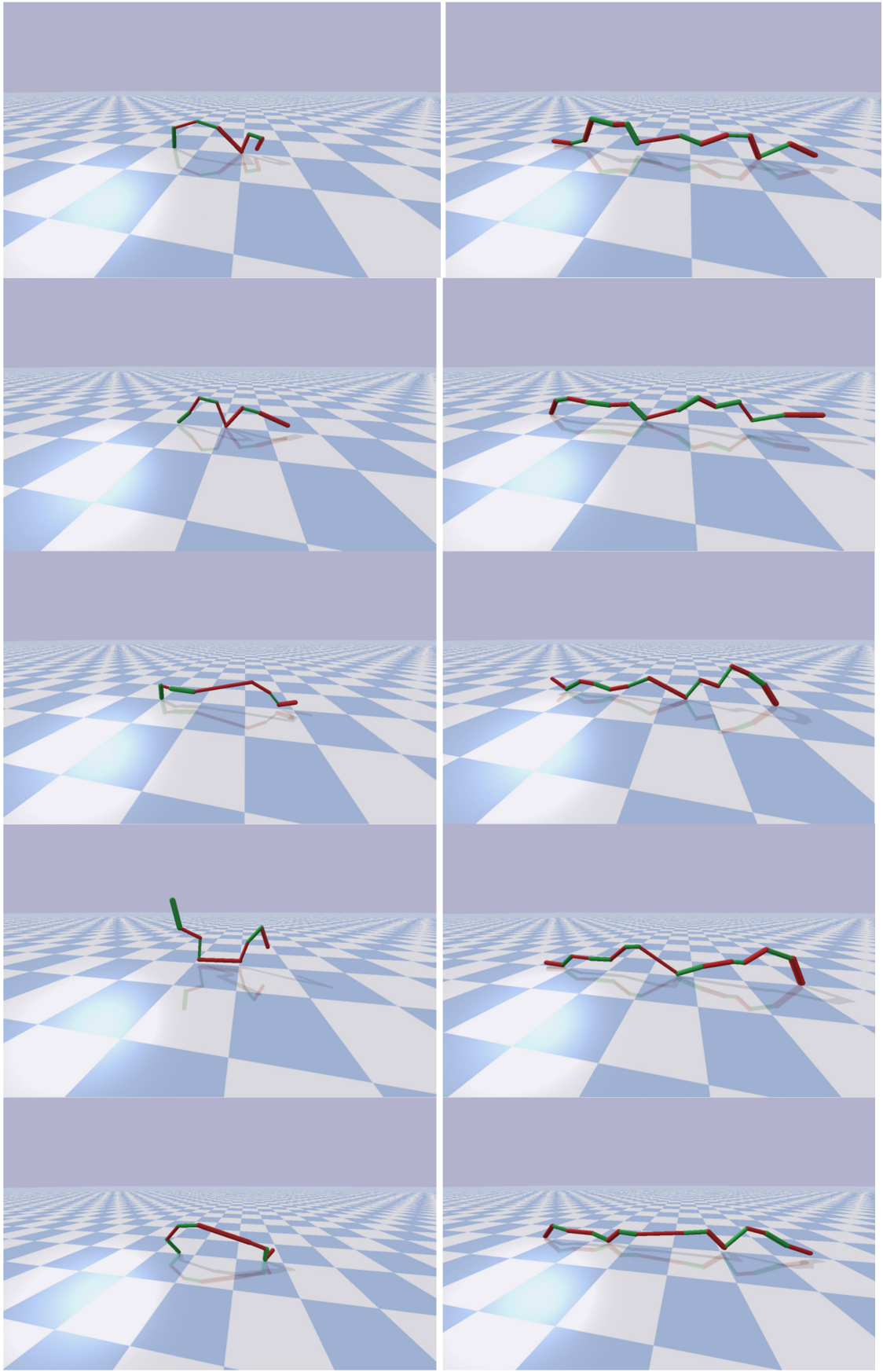

Figure 7. Examples of morphologies evolved in the experiments performed with agents formed by 7 and 13 elements, left and right graph respectively.

Remarkably, also in the case of these experiments, the agents evolved in the coevolving condition outperform the agents evolved in the pre-evolved condition (Figure 6, Mann-Whitney U test with Bonferroni correction, p-value < 0.05). In other words, the agents evolved from scratch outperform the agents possessing the pre-evolved morphology shown in Figure 7. This implies that the advantage provided by the possibility to evolve the morphological characteristics is not simply due to the possibility to discover a suitable morphological bauplan and suitable detailed morphological characteristics. The advantage is due to the possibility to adapt the morphology to the current controller and vice versa to the possibility to adapt the controller to the current morphology. The possession of a suitable morphology, which remains fixed, does not permit to achieve the performance that can potentially be achieved with that morphology. The pre-evolved condition produces much lower performance despite the fact that parts of the traits that should be discovered by evolution are already in place at generation 0.

## 6. Conclusion

Methods capable to co-adapt control and body traits could lead to much better solutions than standard approaches in which the adaptation process is restricted to the control traits only. On the other hand, the methods proposed to date present severe limitations in terms of expressive power and evolvability (Cheney et al., 2016).

In this paper we investigated the efficacy of a direct encoding method combined with a state-of-the-art evolutionary algorithm (Salimans et al., 2017). We showed how this method can be used both to adapt the morphological traits of robots with hand-designed body bauplan and to evolve the bauplan of the robots' body as well, starting from homogeneous morphological structures.

The results obtained by comparing robots with fixed and adapted morphologies confirm that the co-evolution of body and brain leads to significantly better results with respect to standard models in which only the brain is adapted. Interestingly, the advantage is not due to the selection of better morphologies, but rather to the mutual scaffolding process that results from the possibility to co-adapt the morphological traits to the control traits and vice versa. Indeed, robots provided with fixed pre-evolved morphologies did not produce better performance than robot provided with fixed hand-designed morphologies. On the contrary, robots with fixed pre-evolved morphology, that proved effective in the experiments in which the morphology was co-evolved, produced significantly worse performance. The impossibility to co-adapt the morphology to the control system prevented the possibility to discover effective controllers that were instead discovered in the co-evolving experiments.

A second general implication of our results concerns the nature of control and morphological variations and the question of whether morphological variations have destructive effects. The occurrence of a destructive effect has been hypothesized by Cheney et al. (2016) as a consequence of the fact that morphological variations can alter the established communication framework from the controller to the environment. "This result in a system which effectively causes large, unintended variations in the behavior of the controller, as its physical interface is constantly being scrambled while optimization seeks to improve the physical shape of the body" (Cheney et al., 2016, pp. 227).

This destructive effect does not manifest in our experiments. On the contrary, our results indicate that the concurrent variation of the morphological features facilitates the evolution of better solutions. As mentioned above, the concurrent variation of morphological features produces better results than experiments in which the morphological features are pre-adapted and not allowed to vary.

This indicates that whether or not the destructive effect manifests depends on the model used to co-evolve the control and morphological features of the agents. The identification of the detailed characteristics that can generate the disruptive effect might be investigated in future works.

# Appendix

The videos displaying the behavior of one of the best evolved robot with fixed and co-adapted morphology are available at the following links: Halfcheetah: https://youtu.be/pANidyKtWSo, https://youtu.be/SMA2f8B2PVk; Walker2D: https://youtu.be/ro3-rGQ3cwA, https://youtu.be/eox3AbdZCpM. The videos displaying the behavior of one of the best evolved robot with evolved bauplan are available at the following links: https://youtu.be/rSngmzd5hiE and https://youtu.be/dfisG9dn2k0.